\documentclass[10pt,twocolumn,letterpaper]{article}

\usepackage[final]{cvpr}              
\usepackage[dvipsnames]{xcolor}

\definecolor{cvprblue}{rgb}{0.21,0.49,0.74}
\usepackage[pagebackref,breaklinks,colorlinks,citecolor=cvprblue]{hyperref}
\usepackage{multirow}
\usepackage{colortbl}
\usepackage{diagbox}
\usepackage[accsupp]{axessibility} 

\def\paperID{1370} 
\def\confName{CVPR}
\def\confYear{2024}

\title{Benchmarking the Robustness of Temporal Action Detection Models \\ Against Temporal Corruptions}

\author{
Runhao Zeng$^{1,2}$, Xiaoyong Chen$^3$, Jiaming Liang$^3$,
Huisi Wu$^3$, Guangzhong Cao$^{3*}$, Yong Guo$^4$\thanks{Corresponding author}\\
$^1$Artificial Intelligence Research Institute, Shenzhen MSU-BIT University, China,\\
$^2$Guangdong-Hong Kong-Macao Joint Laboratory for Emotional Intelligence and Pervasive Computing,\\
$^3$Shenzhen University
$^4$South China University of Technology}

\begin{document}
\maketitle
\begin{abstract}
Temporal action detection (TAD) aims to locate action positions and recognize action categories in long-term untrimmed videos. Although many methods have achieved promising results, their robustness has not been thoroughly studied. In practice, we observe that temporal information in videos can be occasionally corrupted, such as missing or blurred frames. Interestingly, existing methods often incur a significant performance drop even if only one frame is affected. To formally evaluate the robustness, we establish two temporal corruption robustness benchmarks, namely THUMOS14-C and ActivityNet-v1.3-C. In this paper, we extensively analyze the robustness of seven leading TAD methods and obtain some interesting findings: 1) Existing methods are particularly vulnerable to temporal corruptions, and end-to-end methods are often more susceptible than those with a pre-trained feature extractor; 2) Vulnerability mainly comes from localization error rather than classification error; 3) When corruptions occur in the middle of an action instance, TAD models tend to yield the largest performance drop. Besides building a benchmark, we further develop a simple but effective robust training method to defend against temporal corruptions, through the FrameDrop augmentation and Temporal-Robust Consistency loss. Remarkably, our approach not only improves robustness but also yields promising improvements on clean data. We believe that this study will serve as a benchmark for future research in robust video analysis. Source code and models are available at \url{https://github.com/Alvin-Zeng/temporal-robustness-benchmark}.
\end{abstract}

\begin{figure}[t]
    \centering
    \includegraphics[width=\columnwidth]{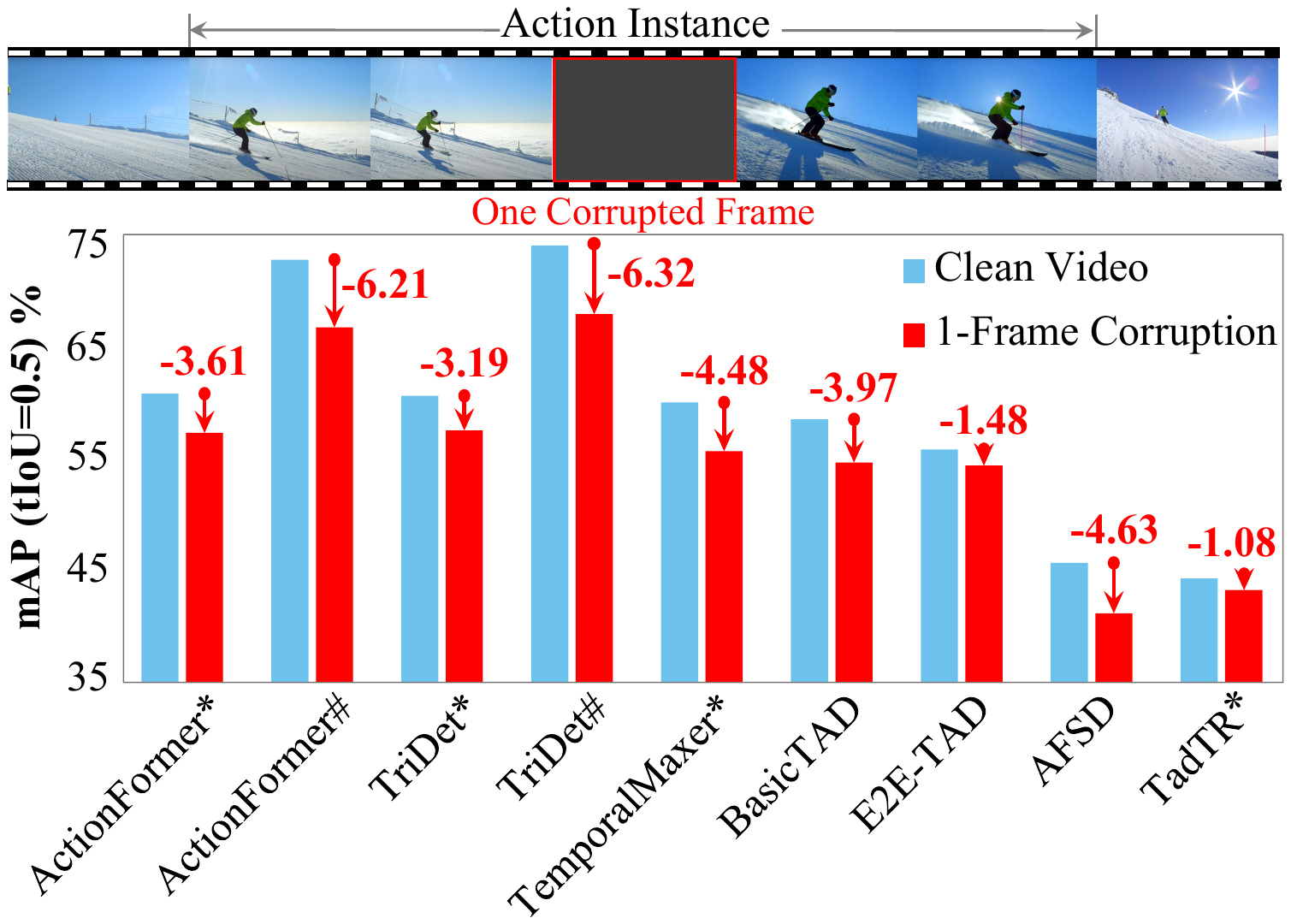}
    \caption{The mAP gap of temporal action detection methods when testing on clean and corrupted videos. * and \# denote the video features extracted by I3D and VideoMAEv2, respectively. Other methods follow an end-to-end manner. Existing TAD methods incur a significant mAP drop of more than 1.08\% even when \textbf{only one frame is corrupted} in an action instance on THUMOS14 dataset, highlighting a prevailing lack of robustness towards temporal corruptions.} 
    \label{Fig.1}
\vspace{-9 pt}
\end{figure}

\section{Introduction}
\label{sec:intro}

Temporal action detection (TAD), an essential aspect of video understanding, seeks to pinpoint action locations and identify action categories in untrimmed videos. Despite the fruitful progress in this mission, the robustness of these methods against corruptions remains largely unexplored. If TAD models are very vulnerable to corruptions, it would become particularly problematic when applying them in various practical contexts, including autonomous driving, security monitoring, and robotics. To verify this, we conduct a preliminary experiment in which we introduce corruptions to a single frame within an action instance, simulating the phenomenon of ``Black Frame''~\cite{choi2015automated} that can occur during data transmission. Remarkably, as shown in Figure~\ref{Fig.1}, the performance of existing TAD methods drops significantly, no matter what kind of features are used or whether the model is trained end-to-end. This outcome reveals that corrupting even a single frame in an action instance disrupts the temporal continuity of the video and damages the temporal information. The poor performance of existing models under these conditions suggests that TAD models generally exhibit weak \textbf{\emph{temporal robustness}}.  As such, a comprehensive evaluation of temporal robustness becomes a cornerstone in advancing this field.



\begin{figure}[t]
    \centering
    \includegraphics[width=\columnwidth]{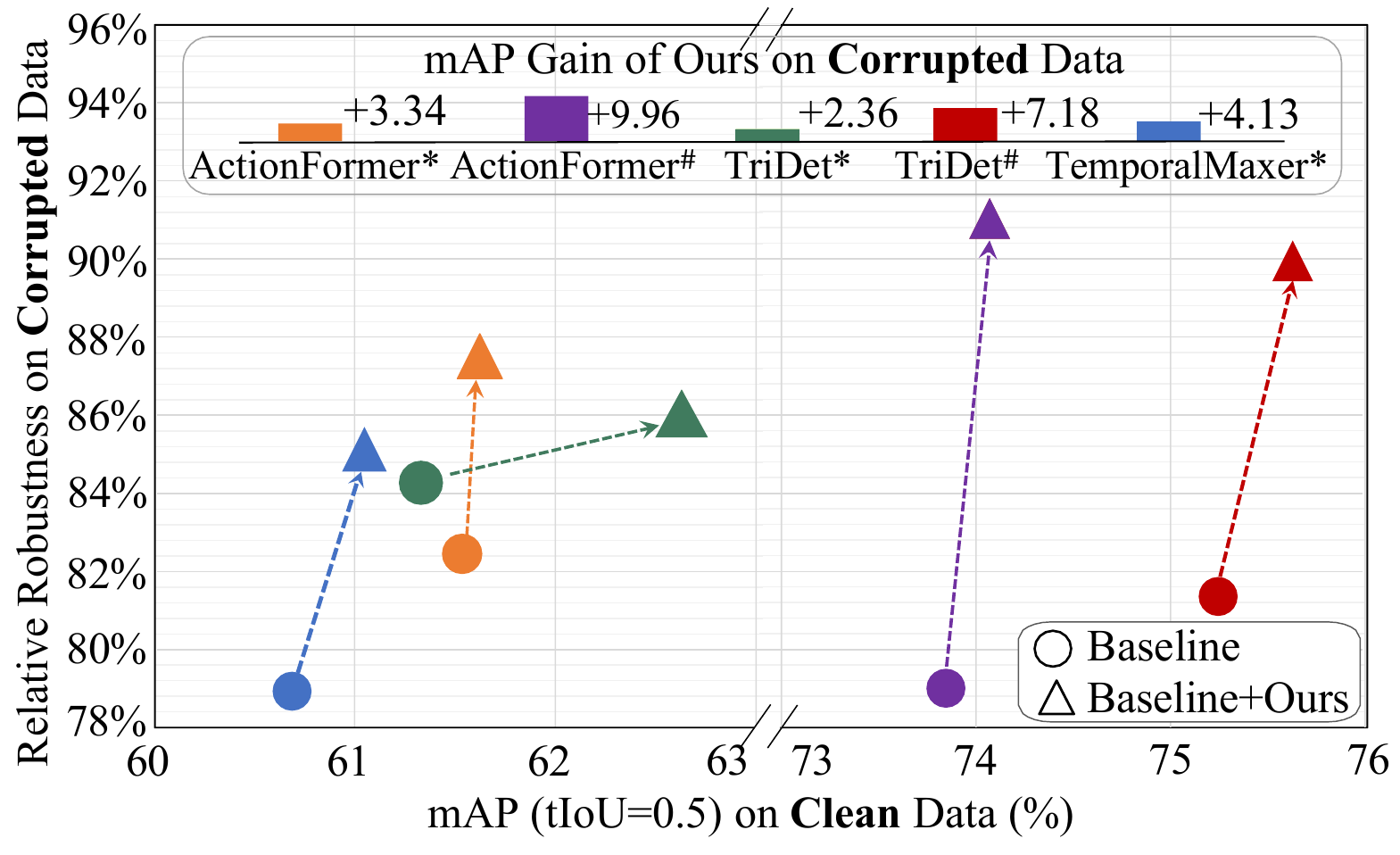}
    \caption{The gain of mAP and relative robustness brought by our proposed training strategy.  * and \# denote the video features extracted by I3D and VideoMAEv2, respectively.
    Our method enhances TAD models' robustness on corrupted videos and, surprisingly, boosts their performance on clean videos.}
    \label{Fig.2} 
\end{figure}

Robustness has been an active research topic yet the majority of studies have concentrated on images \cite{hendrycks2019benchmarking,qin2021understanding}. Recently, in the video domain,~\cite{yi2021benchmarking,schiappa2023large} present robustness benchmarks to evaluate action recognition models.
By contrast, TAD not only requires action recognition; its key distinction lies in the need for temporal localization. Since these benchmarks do not take the change of temporal continuity into account, they are not directly applicable to assess TAD models.
Thus, the design of an effective benchmark capable of evaluating temporal robustness remains an unexplored area.
To this end, we propose two benchmark datasets, THUMOS14-C and ActivityNet-v1.3-C, that contain corruptions in the temporal domain. Specifically, we introduce 5 types of corruptions that are commonly observed in video acquisition and processing. 
To measure the severity of breaking the temporal continuity, we consider 3 levels by varying the number of frames to be corrupted in a video clip.
We conduct in-depth experiments to analyze the robustness based on diverse leading TAD methods and obtain several interesting observations:
1) Existing TAD models demonstrate a notable vulnerability to temporal corruptions. Additionally, it has been observed that end-to-end TAD models are more susceptible to temporal corruptions compared to models that employ a fixed feature extractor. 2) The primary source of this vulnerability can be attributed to localization errors, as opposed to classification errors. 3) The vulnerability of TAD methods is most pronounced when corruption occurs at the center of an action instance. 
We believe these observations may suggest a potential avenue for future research towards robust TAD models.


Furthermore, we also develop a simple but effective method to improve the temporal robustness of TAD models.
First, we propose a \textbf{FrameDrop} augmentation strategy, which randomly selects frames from adjacent actions and backgrounds of a video and introduces corruptions to break the temporal continuity. We highlight that training with such augmented data enables the model to locate and recognize actions against temporal corruptions.
Second, we develop a \textbf{Temporal-Robust Consistency (TRC)} loss, which aligns the model's predictions on corrupted videos with those on clean videos. To increase the efficiency of this alignment, we propose an action-centric sampling strategy, selecting high-quality predictions that are temporally more relevant to the action instance for alignment. Interestingly, our experiments reveal that our robust training method not only increases robustness but also improves performance on clean data (see Figure~\ref{Fig.2}). This study provides essential considerations for future model development.

Our contributions can be summarized as follows:
\begin{itemize}
    \item To the best of our knowledge, we are the first to provide a comprehensive robustness analysis of temporal action detection (TAD) models. We believe that our new observations could be beneficial to developing robust TAD models for real-world deployment.
    \item We build two benchmark datasets and each involves 5 types of corruptions and 3 severities, resulting in 15 corruption types in total. We show that existing TAD methods are very vulnerable and often incur a significant performance drop on our benchmarks. Besides the performance on clean data, we highly recommend that researchers additionally evaluate their models in terms of temporal robustness in future research.
    \item We propose a simple but effective training method to improve temporal robustness. Interestingly, our method not only improves the robustness based on a diverse set of popular TAD models on corrupted videos but also obtains better performance on clean data in most cases.
\end{itemize}

\section{Related Work}
\label{sec:related_work}

\subsection{Temporal Action Detection}
Temporal action detection approaches can be grouped into two categories: \textbf{Two-stage methods} primarily involves generating a set of proposals followed by their classification and boundary refinement~\cite{shou2016temporal,xu2017r,zhao2017temporal,zeng2021graph,chao2018rethinking}. To generate proposals, one can perform frame or segment-level classification and merge frames or segments of the same category~\cite{shou2017cdc,montes2016temporal,piergiovanni2019temporal,nag2022proposal}, while other methods use proposal generation methods~\cite{lin2019bmn,buch2017sst}. However, these methods heavily depend on the quality of the proposals, leading to the development of integrated approaches that combine proposal generation with classification and/or boundary regression~\cite{yeung2016end,buch2017end,lin2017single,huang2019decoupling,tang2023temporalmaxer}, referred to \textbf{One-stage methods}. Notable contributions include the introduction of the anchor mechanism for TAD by~\cite{lin2017single} and the exploration of anchor-free schemes~\cite{shi2023tridet,lin2021learning}, further advanced the field by merging the advantages of both anchor-based and anchor-free methods~\cite{yang2020revisiting}. Recently, transformer-based models, which have shown remarkable success in various vision tasks, have been adapted to TAD~\cite{zhang2022actionformer}. Other advances like Graph Convolution are also introduced to this task~\cite{xu2020gtad,zeng2019graph} and end-to-end architectures have been explored in~\cite{liu2022end,yang2023basictad,liu2022empirical}. Despite their success, they
focused on training and testing on a benchmark dataset with little distribution shift from training to testing samples, which poses challenges for real-world applications. This paper aims to investigate the robustness of TAD models and enhance their performance, particularly under conditions of corruptions.

\subsection{Robustness of Neural Networks}

\textbf{Image Domain}.
Despite the outstanding performance of deep neural networks, they are not robust to image corruptions~\cite{hendrycks2021many}. 
To address this, recent work explores re-calibrating batch normalization statistics \cite{SchneiderARXIV2020,BenzARXIV2020,NadoARXIV2020} or utilizing the frequency domain \cite{SaikiaARXIV2021} to improve corruption robustness.
However, data augmentation methods such as \cite{RusakECCV2020,geirhos2018imagenet,cubuk2018autoaugment,hendrycks2019augmix,hendrycks2021many} represent the most prominent and successful line of work, ranging from simple Gaussian noise augmentation \cite{RusakECCV2020}, over well-known schemes such as AutoAugment \cite{cubuk2018autoaugment} to strategies specifically targeted towards corruption robustness such as AugMix \cite{hendrycks2019augmix} or DeepAugment \cite{hendrycks2021many}.
Besides random corruptions, deep networks are susceptible to adversarial examples \cite{SzegedyARXIV2013,GoodfellowARXIV2014}.
While plenty of approaches for defending against adversarial examples have been proposed~\cite{SilvaARXIV2020,BarrenoASIACCS2006,YuanARXIV2017,AkhtarARXIV2018,BiggioARXIV2018,XuARXIV2019,ChaubeyARXIV2020,PangICLR2021,GowalARXIV2020}, adversarial training (AT) has become the de facto standard \cite{madry2017towards}.
On the other hand, many efforts have been made to design or train a robust architecture to improve model robustness.
Due to the success of Vision Transformers (ViTs)~\cite{DosovitskiyICLR2021,VaswaniNIPS2017,heo2021rethinking}, many works seek to study and improve the robustness of ViTs~\cite{mao2021towards,bhojanapalli2021understanding,bai2021transformers,shi2020robustness,guo2023improving,paul2022vision,benz2021robustness,han2022robustify,guo2023robustifying,zhou2022understanding}.
Interestingly, ViTs are often more robust than convolutional networks against corruptions~\cite{qin2021understanding,TangARXIV2021,tian2022deeper} and adversarial attacks~\cite{MahmoodICCV2021,shao2021adversarial,naseer2021improving,lovisotto2022give,liang2022not,benz2021adversarial,tang2022exploring,gu2022evaluating}. Nevertheless, most of them mainly focus on the robustness issue in the image domain, leaving the robustness in the temporal domain (\eg, inside videos) unexplored.

\begin{figure*}[t]
    \includegraphics[width=\textwidth]{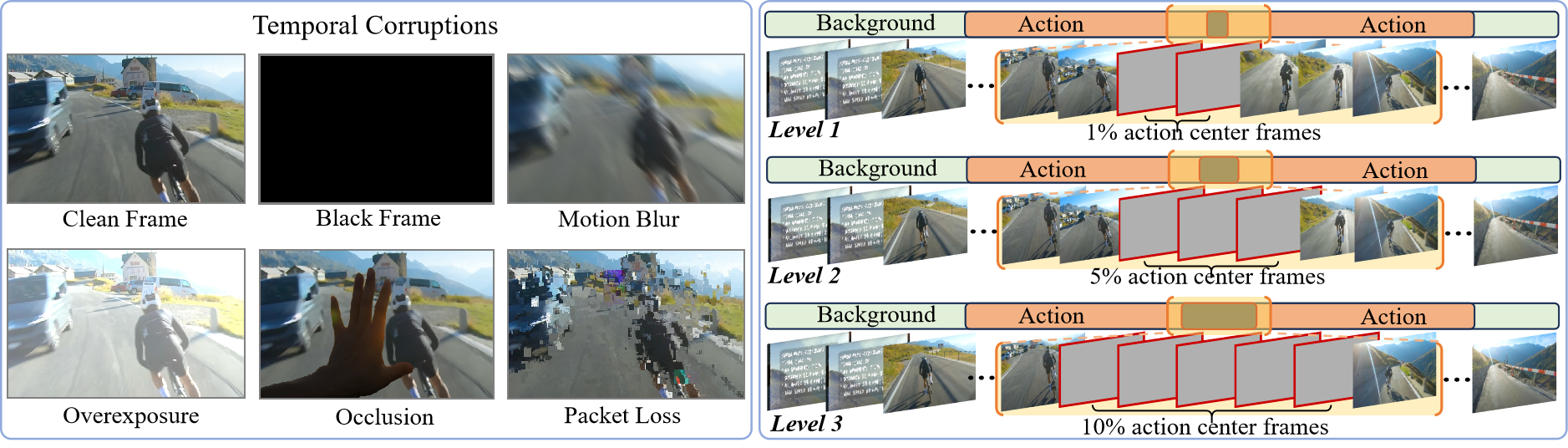}
    \caption{Our temporal robustness study introduces 5 types of temporal corruptions that are frequently encountered in real-world scenarios, including black frame~\cite{choi2015automated}, motion blur, overexposure, occlusion and packet loss~\cite{yi2021benchmarking}. Each type of corruptions has 3 levels of severity and each level refers to the $l\%$ ($l \in \{1,5,10\}$) action center frames being corrupted, eventually resulting in 15 distinct corruptions.}
    \label{Fig.3_dataset}
\end{figure*}

\noindent \textbf{Video Domain}. 
Data augmentation techniques have been shown effective in improving the robustness of video analysis models. Li \etal \cite{li2019dynamic} utilized temporal cropping as video data augmentation. 
Mixup, cutmix, and cutout operations from the image domain were introduced to the video domain~\cite{kim2020learning,yun2020videomix,kim2022exploring}. Isobe \etal \cite{isobe2020intra} proposed the application of the same transformation across all frames in each mini-batch of video clips. In another line of research, Zhang \etal \cite{zhang2020self} used a Generative Adversarial Network (GAN) to create dynamic images that encapsulate motion information from video. Wu \etal \cite{wu2019adversarial} developed a generator to produce a frame encompassing all motion feature information. The robustness of video models against common corruptions has recently been analyzed \cite{yi2021benchmarking,schiappa2023large}. 
They benchmark the robustness of common convolutional- and transformer-based spatio-temporal architectures, against several corruptions in video acquisition and video processing. The corruptions they used can be generated across a set of continuous frames or
depend solely on the content of a single frame. Their benchmarks apply corruptions to all frames of a trimmed video, aiming to measure the robustness of action recognition models. In this paper, we are particularly interested in examining the temporal detection robustness of TAD models and creating benchmarks by applying corruptions to a subset of the video frames to disrupt the temporal continuity.




\section{Temporal Robustness Benchmark Creation}
\subsection{TAD Formulation and Notation} \label{Sec:notation}
Temporal action detection (TAD) requires a machine to recognize the action instances and simultaneously identify their temporal positions in a video. Given an untrimmed video as $V=\{I_t\}_{t=1}^T$, where $I_t$ denotes the frame at the time slot $t$, TAD predicts a set of action instances $\Phi_{V}=\{\phi_i=(t_{s_i},t_{e_i},k_{i}) \}_{{i}=1}^{N}$, where $N$ is the number of action instances in $V$, $t_{s_i}$, $t_{e_i}$ and $k_{i}$ are the starting time, ending time, and category of the $i$-th action instance, respectively. 

\subsection{Temporal Corruptions in Videos}

Our study begins with a scenario commonly encountered in daily life. We observed that during video playback, certain frames may suddenly experience interference and immediately disappear. For instance, an object might abruptly enter and then exit the frame during recording, or sudden changes in lighting might cause overexposure, which is then corrected by the camera. For humans, such interferences have minimal impact on our ability to locate actions in a video. However, our preliminary experiments (see Figure~\ref{Fig.1}) demonstrate that even a single corrupted frame in an action sequence can significantly impair the temporal localization performance of TAD models.

Consequently, our research focuses on temporal corruptions that appear abruptly and vanish just as quickly during video recording—a common phenomenon in videos but one not previously addressed in research. Formally, given a video $V=\{I_1,I_2,\dots,I_t,\dots,I_T\}$, our approach to corruptions does not encompass all frames. Instead, we replace specific clean frames with corrupted ones, resulting in a corrupted video $V^c=\{I_1,I_2,\dots,I^c_t,\dots,I_T\}$ to disrupt the temporal continuity, where $I^c_t$ represents the corrupted version of $I_t$. We study five categories of real-world corruptions, as depicted in Figure~\ref{Fig.3_dataset}, including:
\begin{itemize}
    \item \textbf{Black frame}~\cite{choi2015automated}: caused by transferring tape-based content to digital files or temporary network disconnections during video streaming
    \item \textbf{Motion blur}~\cite{hendrycks2019benchmarking}: occurs when the camera undergoes swift and rapid movements
    \item \textbf{Overexposure}~\cite{hendrycks2019benchmarking}: due to fluctuations in daylight intensity, or sudden changes in photographic conditions
    \item \textbf{Occlusion}: resulting from the camera being accidentally blocked by another object while filming
    \item \textbf{Packet loss}~\cite{yi2021benchmarking}: arising from video transmission over imperfect channels in real-world settings
\end{itemize}
\noindent \textbf{Discussion:} Existing video robustness benchmarks apply corruptions across all frames, such an approach does not effectively validate temporal localization performance. With TAD models requiring both localization and recognition, it is challenging to ascertain whether issues arise in localization or recognition. By defining temporal corruptions where the majority of frames in a video remain clean, the impact on recognition is minimized. However, the few corrupted frames we introduce, although limited in number, directly disrupt the temporal continuity. Experiments have validated that our proposed corruption method effectively tests the localization capabilities of TAD models (see Section \ref{Sec:localization_error}).

\subsection{Benchmark Datasets}
To benchmark the robustness of the TAD models against the temporal corruptions, we create two benchmark datasets, including THUMOS14-C and ActivityNe-v1.3-C.


\noindent \textbf{THUMSO14-C.} As a standard benchmark for action detection, THUMOS14~\cite{jiang2014thumos} contains a training set, known as the UCF-101 dataset, which consists of 13320 videos. The validation, testing, and background sets contain 1010, 1574, and 2500 untrimmed videos, respectively. The temporal action detection task of THUMOS14, which contains videos over 20 hours from 20 sports classes, is very challenging since each video has more than 15 action instances and its 71\% frames are occupied by background items. In this study, we apply 5 distinct corruptions, each at 3 levels of severity, to the 213 annotated videos from the testing set. Specifically, we introduce corruptions to the central $l\%$ of frames in each action instance, where $l \in \{1,5,10\}$. Level 1 indicates a minimal temporal corruption that affects only 1\% of the frames within a given action instance while Level 3 signifies a more substantial temporal corruption that affects 10\% of the frames. We choose to corrupt the central frames since we empirically found that the robustness of TAD models degrades more significantly when the corrupted frame is located closer to the center of an action instance (see Section~\ref{Sec:ablated_center} for more results).

\noindent \textbf{ActivityNet-v1.3-C.} ActivityNet~\cite{caba2015activitynet} is a popular benchmark for TAD on untrimmed videos. We create a benchmark on ActivityNet-v1.3, which contains approximately 10K training videos and 5K validation videos corresponding to 200 different activities. Each video has an average of 1.65 action instances. Similarly, we apply the proposed 5 corruptions with 3 severity levels to the validation set.



\subsection{Robustness Metrics}

We first introduce the standard mean Average Precision (mAP) that is widely used to evaluate TAD models. We further take the mAP on clean data into account and develop a new metric to measure how large the performance drop would be between the mAP on clean and corrupted data.

\noindent \textbf{Mean Average Precision (mAP)} is a commonly used evaluation metric for action detection performance. A predicted temporal bounding box is considered to be correct if its temporal IoU with the ground-truth instance is larger than a certain threshold and the predicted category is the same as this ground-truth instance. On THUMOS14-C, the tIOU thresholds are chosen from $\{0.1, 0.2, 0.3, 0.4, 0.5\}$ and we report mAP@tIoU=0.5 for comparisons; on ActivityNet-v1.3-C, the tIoU thresholds are from $\{0.5, 0.75, 0.95\}$, and we report the average mAP of the tIoU thresholds between 0.5 and 0.95 with the step of $0.05$.

\noindent \textbf{Relative robustness}. We introduce a new metric, termed as relative robustness $\gamma^r$ to measure the robustness of TAD models. We first calculate the mAP $M_{clean}$ on the clean test set given a trained model $g$. Then, we test $g$ on a corruption $c$ at each of the severity levels $s$, and obtain mAP $M_{c,s}$. It should be noted that different models exhibit diverse performances on identical test videos, thus, an absolute drop in performance is also influenced by the model's performance on clean videos. Therefore, we determine relative performance drop as a measure of the model's robustness. Each severity level $s$ and corruption $c$ has its own relative robustness $\gamma^r_{c,s}$ computed as $\gamma^r_{c,s} = 1 - (M_{clean} - M_{c,s})/M_{clean}$. We average across all severity levels and corruptions to yield to yield $\gamma^r$ of a TAD model.

\section{Benchmarking Robustness of TAD Models}
In our benchmark study, we train the TAD models with clean data and evaluate them on the corrupted data. It is a standard setting under the robust generalization study \cite{YinNIPS2019}, which assumes that the model is unable to know the exact problem in the deployment in advance.

\noindent \textbf{Model Variants.} 
Our experiments evaluate seven popular TAD models, employing various architectural frameworks such as CNN, Transformer, and Graph Convolution. Regarding the detection heads, ActionFormer~\cite{zhang2022actionformer} and E2E-TAD~\cite{liu2022empirical} employs a Transformer architecture, while TriDet~\cite{shi2023tridet}, AFSD~\cite{lin2021learning}, BasicTAD~\cite{yang2023basictad}, and TemporalMaxer~\cite{tang2023temporalmaxer} are CNN-based models. VSGN~\cite{zhao2021video} is constructed based on graph architectures. Note that E2E-TAD, BasicTAD and AFSD are trained in an end-to-end manner while others rely on pre-trained feature extractors. We exploit three different feature extractors, including  I3D~\cite{carreira2017quo}, VideoMAEv2~\cite{wang2023videomae}, and TSP~\cite{alwassel2021tsp}. Among these, the I3D model leverages 3D convolutions, whereas VideoMAEv2 adopts a Transformer-based approach, trained using a dual masking strategy. TSP, on the other hand, utilizes a ResNet-based backbone pre-trained on temporal sensitive tasks. In our experimental setup, we use the official implementations and pre-trained weights of these models.





\begin{table}[t]
\tabcolsep 4pt
\centering
\caption{Corruption robustness of TAD models on THUMOS14-C. $^*$ denotes end-to-end methods. Existing TAD models are particularly vulnerable to temporal corruptions, regardless of whether they are based on transformers or CNN.}
\resizebox{\columnwidth}{!}{
\begin{tabular}{l|l|c|c|c}
\hline
\multirow{2}{*}{Model} & \multirow{2}{*}{Feature} & \multirow{2}{*}{\shortstack{Clean \\ mAP}} & \multirow{2}{*}{\shortstack{Corrupted \\ mAP}}  & \multirow{2}{*}{\shortstack{Relative \\ Robustness}} \\
 & & & & \\
\hline
BasicTAD$^*$~\cite{yang2023basictad} & SlowOnly & 59.17 & 37.72 (21.45 $\downarrow$) & 63.75 \\
E2E-TAD$^*$~\cite{liu2022empirical} & SlowFast & 56.41 & 30.55 (25.86 $\downarrow$) & 54.16 \\
TemporalMaxer~\cite{tang2023temporalmaxer} & I3D & 60.72 & 47.82 (12.90 $\downarrow$) & 78.76 \\
ActionFormer~\cite{zhang2022actionformer} & I3D & 61.53 & 50.61 (10.92 $\downarrow$) & 82.25\\
ActionFormer~\cite{zhang2022actionformer} & VideoMAEv2 & 73.84 & 58.33 (15.51 $\downarrow$) & 78.99 \\
AFSD$^*$  ~\cite{lin2021learning} & I3D & 46.05 & 34.47 (11.58 $\downarrow$) & 74.85 \\
TriDet~\cite{shi2023tridet} & I3D & 61.33 & 51.71 (9.62 $\downarrow$) & 84.31 \\
TriDet~\cite{shi2023tridet} & VideoMAEv2 & 75.16 & 61.10 (14.06 $\downarrow$) & 81.29 \\
\hline
TriDet~\cite{shi2023tridet}+Ours & VideoMAEv2 & \textbf{75.60} & \textbf{68.28 (7.32$\downarrow$)} & \textbf{90.31} \\
\hline
\end{tabular} 
}
\label{Tab:THUMOS14-C}
\end{table}

\begin{table}[t]
\tabcolsep 3pt
\centering
\caption{Corruption robustness of TAD models on ActivityNet-v1.3-C. $^*$ denotes end-to-end methods. TAD models remains highly susceptible to temporal corruptions, suggesting that the vulnerability is not specific to any particular dataset.}
\resizebox{\columnwidth}{!}{
\begin{tabular}{l|l|c|c|c}
\hline
 \multirow{2}{*}{Model} & \multirow{2}{*}{Feature} & \multirow{2}{*}{\shortstack{Clean \\ mAP}} & \multirow{2}{*}{\shortstack{Corrupted \\ mAP}}  & \multirow{2}{*}{\shortstack{Relative \\ Robustness}} \\
 & & & & \\
\hline
VSGN~\cite{zhao2021video} & I3D & 31.85 & 30.08 (1.77 $\downarrow$)& 94.44 \\
TriDet~\cite{shi2023tridet} & TSP & 36.66 &  15.18 (21.48 $\downarrow$)& 41.41 \\
ActionFormer~\cite{zhang2022actionformer} & TSP & 36.50 & 27.79 (8.71 $\downarrow$) & 76.12 \\
ActionFormer~\cite{zhang2022actionformer} & VideoMAEv2 & 38.47 & 33.93 (4.54 $\downarrow$)& 88.19 \\ 
AFSD$^*$ ~\cite{lin2021learning} & I3D & 32.49 & 29.56 (2.93 $\downarrow$) & 90.98  \\ 
\hline
AFSD$^*$ ~\cite{lin2021learning}+Ours & I3D & \textbf{32.86} & \textbf{30.78 (2.08 $\downarrow$)} & \textbf{93.68}  \\ 
\hline
\end{tabular}
}
\label{Tab:Anet-C}
\end{table}


\subsection{Existing TAD Models are Particularly Vulnerable to Temporal Corruptions}

The robustness against temporal corruptions of TAD models on the THUMOS14-C dataset is presented in Table~\ref{Tab:THUMOS14-C}.
All TAD models assessed in this study show susceptibility to temporal corruptions, evidenced by a reduction in detection mAP ranging from 9.62\% to 25.86\%. This vulnerability is observed regardless of the type of features employed or whether the model is end-to-end trained.
When considering the relative robustness calculated as an average across five types of corruptions and three levels, the highest-performing model, Tridet, achieves only 84.31\%, while the lowest, E2E-TAD, scores 54.16\%.  This indicates that the robustness of TAD models is influenced by both the model architecture and the input data characteristics. Furthermore, we observe that methods employing end-to-end training, such as BasicTAD, E2E-TAD, and AFSD, are more susceptible to temporal corruptions compared to those utilizing a fixed feature extractor approach (when using the same backbone).
We also compare the temporal robustness of existing TAD models on ActivityNet-v1.3-C and report the results in Table~\ref{Tab:Anet-C}. On this dataset, the performance of TAD models remains highly susceptible to temporal corruptions. The range of decrease in terms of mAP spans from 1.77\% to 21.48\%. This suggests that the limited temporal robustness of TAD models is not specific to any particular dataset. Please refer to the supplementary material for detailed performance on each type of corruption at every level.

\begin{figure}[t]
    \centering
    \includegraphics[width=\columnwidth]{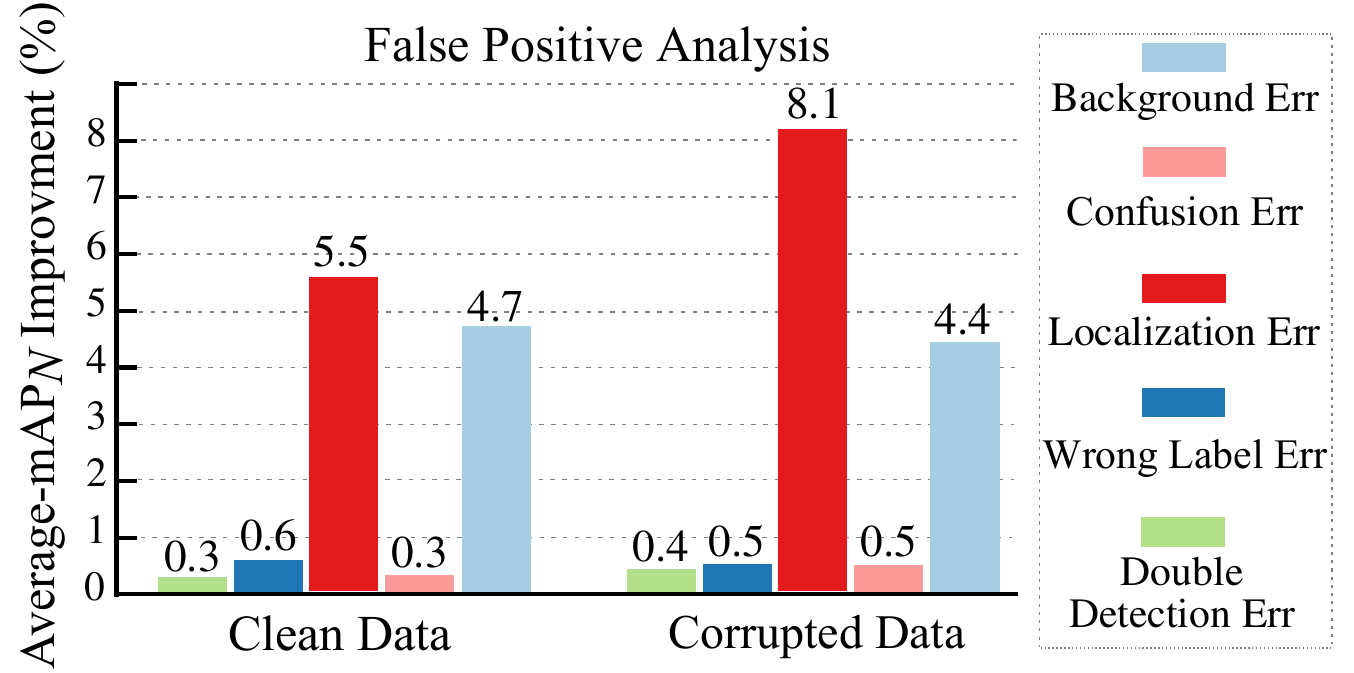}
    \caption{False positive profiling of the TriDet's predictions on THUMOS14-C. The Wrong Label (classification) Error remains relatively consistent, whereas the Localization Error increases significantly on corrupted data, revealing that vulnerability mainly comes from localization error rather than classification error.}
    \label{Fig.false_positive}
\end{figure}

\subsection{Vulnerability Mainly Comes from Localization Error rather than Classification Error}\label{Sec:localization_error}
We follow DETAD~\cite{alwassel2018diagnosing} to analyze the results predicted by the TriDet model. We categorize the causes leading to false positives predicted by the model into five types, including 1) Background Error: predict action as background, 2) Confusion Error: low-quality boundary with wrong category, 3) Localization Error: low-quality boundary with correct category, 4) Wrong Label Error: high-quality boundary with correct category, 5) Double Detection Error: two predictions match one action. Figure~\ref{Fig.false_positive} illustrates the enhancement in the performance of the model when a certain type of error is eliminated. It is evident that the impact of Localization Error escalates from 5.5\% to 8.1\% upon the corrupted video, while the Wrong Label Error (0.6\% v.s. 0.5\%) remains relatively stable, indicating a minor influence on classification. This suggests that our proposed benchmark primarily assesses the robustness of the model's temporal localization in the face of temporal corruptions. Compared to other benchmarks designed for action recognition robustness ~\cite{yi2021benchmarking,schiappa2023large}, ours is more suitable to examine the temporal robustness of TAD models. Please kindly refer to the supplementary material for more analysis of other TAD models.

\subsection{Corrupting the Central Frames of Action Results in the Strongest Attack}\label{Sec:ablated_center}
\begin{figure}[t]
    \centering
    \includegraphics[width=\columnwidth]{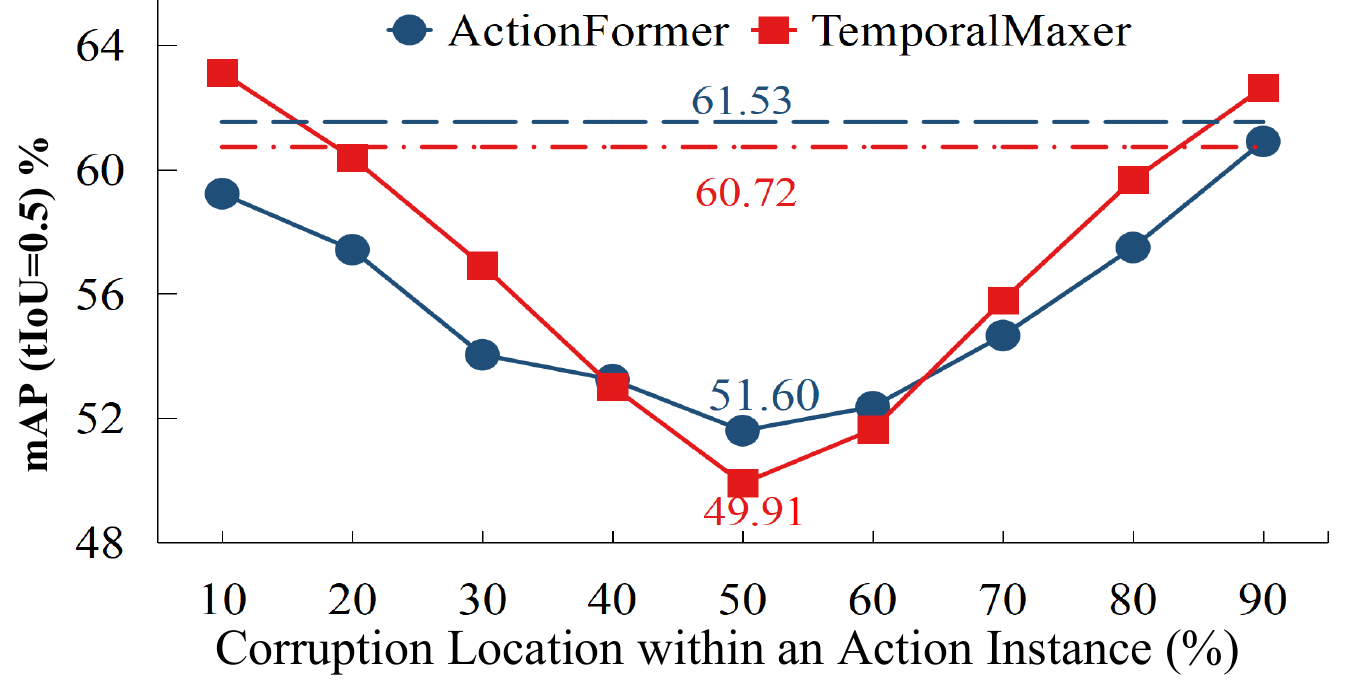}
    \caption{The performance of TAD models with varying corruption locations within an action instance on THUMOS14-C. The horizontal dashed lines refer to the model's performance on clean videos. As corruptions approach the center, its impact on the model becomes increasingly significant.}
    \label{fig:drop_position}
\end{figure}

To investigate the impact of corruption's location within action instances, we evaluate the robustness of ActionFormer and TemporalMaxer on the THUMOS14-C dataset. We sample five continuous frames at every 10\%, 20\%, \dots, and 90\% of each action instance and introduce black frame corruption. From Figure~\ref{fig:drop_position}, when corruption is centered within an action instance, both models exhibit the most pronounced performance degradation. Interestingly, we also discover that the models' performance improved when corruptions are introduced near the boundaries of the action, surpassing the performance of clean data. We posit that this is due to the models interpreting the position of the black frame as the boundary of the action, thus improving the localization performance. Consequently, we opt to replace the frames at the center of each action instance with corrupted frames, in order to construct temporal corrupted datasets that maximally disrupt model performance.

\begin{figure*}[t]
    \centering 
    \includegraphics[width=\textwidth]{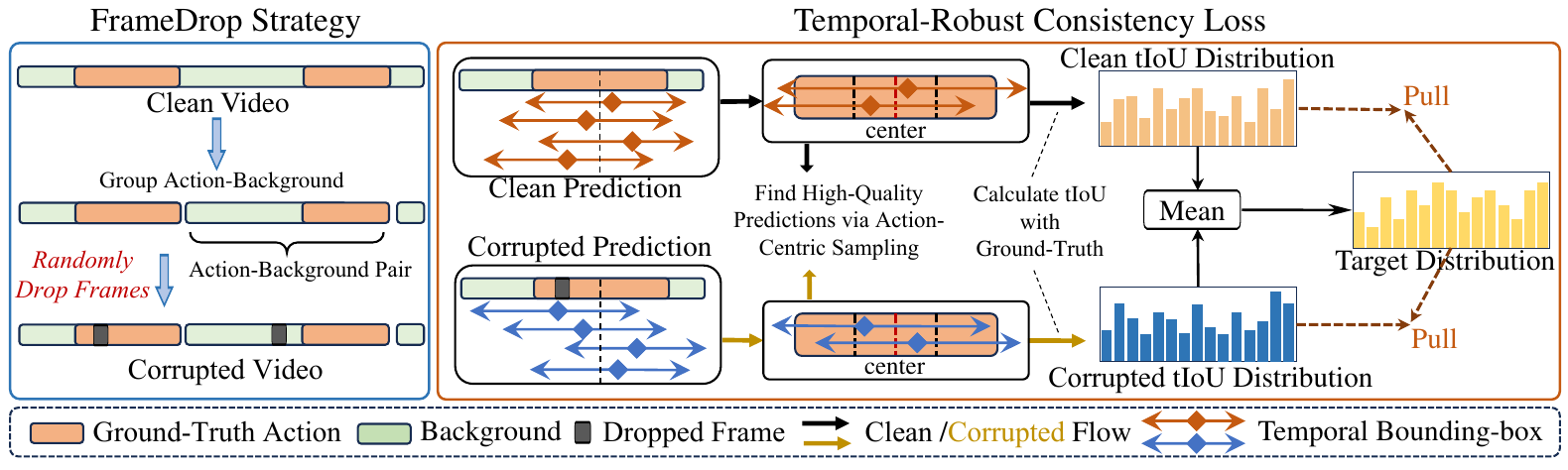}
    \caption{Our approach towards training a robust TAD model. We propose a FrameDrop strategy to interrupt the temporal continuity by first grouping the adjacent action and background instance into pairs and randomly replacing a clean frame within each pair with a black frame. Then, we perform action-centric sampling to find high-quality predictions from clean and corrupted videos and propose a temporal-robust consistency loss to align them in terms of their tIoU distributions w.r.t ground-truth actions.}
    \label{Fig.4_loss} 
\end{figure*}

\section{Defending against Temporal Corruptions}\label{Sec:method}


With the above observations, we seek to improve existing TAD models' temporal robustness from two perspectives. First, in Section~\ref{Sec:FD}, we propose a \textbf{FrameDrop strategy} to simulate temporal corruptions. Our intuition is that during model training, introducing corruptions to the input video forces the model to better leverage the uncorrupted temporal context for localizing action and identifying action categories. Second, in Section~\ref{Sec:TRC}, we develop a new \textbf{Temporal-Robust Consistency (TRC) loss} to improve the localization capability by guiding the model to predict temporal bounding boxes that are temporally related to actions.


\subsection{FrameDrop Strategy}\label{Sec:FD}
 We propose to randomly drop frames from the input video to corrupt the temporal continuity. In particular, we divide the input video into multiple Action-Background (AB) pairs (\ie, $V=\{P^{AB}_i\}_{i=1}^{N_p}$), where each pair is composed of adjacent action and background segments and $N_p$ is the number of the resultant pairs. Then, within each AB pair $P^{AB}=\{\ I^a_1,I^a_2,\dots,I^a_{N_a},I^b_1,I^b_2,\dots,I^b_{N_b}\}$, where $N_a$ and $N_b$ refer to the number of frames in the action and background segment, we randomly select a single frame to drop (\ie, replace with a black frame). The video, once subjected to FrameDrop, is then forwarded to a specific feature extractor (\eg, I3D, VideoMAEv2) or directly fed into the TAD model (\eg, the end-to-end method BasicTAD). 

It is noteworthy that our method performs FrameDrop operations in both action and background segments. This prevents the model from memorizing that corruptions will occur in the action segments, thus circumventing a trivial solution. Further experiments are detailed in the supplementary materials. We empirically show that our FrameDrop is able to consistently improve the robustness of various TAD models against different corruptions that are unseen during training (see Section~\ref{Sec:improvement} for more results).

\subsection{Temporal-Robust Consistency Loss}\label{Sec:TRC}

When both corruppted videos and clean videos processed through FrameDrop are fed into the TAD model, two sets of temporal bounding box $\hat{\Phi}^c=\{\phi_i^c=(t_{s_i}^c,t_{e_i}^c,k_{i}^c) \}_{{i}=1}^{N_c}$ and $\hat{\Phi}^d=\{\phi_i^d=(t_{s_i}^d,t_{e_i}^d,k_{i}^d) \}_{{i}=1}^{N_d}$ are predicted, where $N_c$ and $N_d$ are the number of predictions. A simple training approach would be to ensure consistency between these two prediction sets. However, this strategy presents two issues. The first is computational efficiency, as a single video can contain numerous predictions. More critically, the TAD method's primary focus is the localization of actions, hence our attention should be primarily focused on predictions temporally related to action instances. As a result, we propose an action-centric sampling strategy for selecting action-related predictions. 

Without loss of generality, let a ground-truth action instance be $\phi=(t_{s},t_{e},k)$, where we omit $i$ for simplicity. We compute the timestamp of its central frame by $t^* = \frac{t_{e}+t_{s}}{2}$ as a reference. Similarly, for the predicted temporal bounding boxes in $\hat{\Phi}^c$ and $\hat{\Phi}^d$, we could also calculate the timestamp of their central frame and obtain two central location sets $T^c=\{t^c_j\}^{N_c}_{j=1}$ and $T^d=\{t^d_j\}^{N_d}_{j=1}$, respectively.
Then, we choose the top-K predictions from $T^c$ (or $T^d$) with the minimal distance w.r.t $t^*$ and calculate the temporal Intersection over Union (tIoU) between the selected $K$ predictions and the ground truth (GT) action instance $\phi$. Thus, we obtain two tIoU distributions - one corresponding to the corrupted input $p_c\in \mathbb{R}^K$, and the other to the clean input $p_d\in \mathbb{R}^K$. To align the predictions under these two circumstances, we average $p_c$ and $p_d$ to serve as the target tIoU distribution $p_t$, and then separately calculate the Kullback-Leibler (KL) divergence of the predicted distribution w.r.t $p_t$. In this way, we derive the temporal-robust consistency loss
\begin{equation}
    L_{TRC} = \frac{1}{2}(\mathrm{KL}[p_t||p_c]+\mathrm{KL}[p_t||p_d]).
\end{equation}
Note that we could instead compute the KL divergence between $p_c$ and $p_d$ or adopt the Mean Squared Error (MSE) as the loss function, but these do not perform as well. More ablated results of the loss function can be found in the supplementary material.

\section{Improved Robustness and Further Analysis}





\begin{table}[t]
\tabcolsep 3pt
\centering
\caption{Results of defending against temporal corruptions with the help of our proposed training strategy on THUMOS14-C. Our method consistently improves the robustness of various TAD models with different features.} 
\resizebox{\linewidth}{!}{
\begin{tabular}{l|l|l|l}
\hline
\multirow{2}{*}{Backbone(feature)} & \multirow{2}{*}{\shortstack{Clean \\ mAP}} & \multirow{2}{*}{\shortstack{Corrupted \\ mAP}}  & \multirow{2}{*}{\shortstack{Relative \\ Robutness}}\\ 
& & & \\
\hline
TemporalMaxer(I3D) & 60.72 & 47.82 & 78.76 \\
 + Ours & \textbf{61.04 (0.32 $\uparrow$)}& \textbf{51.95 (4.13 $\uparrow$)} & \textbf{85.10 (6.34 $\uparrow$)} \\  \hline
TriDet(I3D) & 61.33 & 51.71  & 84.31 \\
  + Ours & \textbf{62.63 (1.30 $\uparrow$)}& \textbf{54.07 (2.36 $\uparrow$)} & \textbf{86.32 (2.01 $\uparrow$)} \\ \hline
TriDet(VideoMAEv2) & 75.16 & 61.10 & 81.29 \\
 + Ours & \textbf{75.60 (0.44 $\uparrow$)}& \textbf{68.28 (7.18 $\uparrow$)} & \textbf{90.32 (9.03 $\uparrow$)} \\ \hline
ActionFormer(I3D) & 61.53 & 50.61 & 82.25 \\
 + Ours & \textbf{61.63 (0.10 $\uparrow$)}& \textbf{53.95 (3.34 $\uparrow$)}  & \textbf{87.54 (5.29 $\uparrow$)} \\  \hline
ActionFormer(VideoMAEv2) & 73.84 & 58.33  & 78.99 \\
 + Ours & \textbf{74.06 (0.22 $\uparrow$)}& \textbf{68.29 (9.96 $\uparrow$)}  & \textbf{92.21 (13.22 $\uparrow$)} \\ \hline

\end{tabular}
}\label{Tab:THUMOS14_Improve}
\end{table}


\subsection{Improved Robustness of TAD Models}
In addressing the performance drop caused by temporal corruptions, we propose to enhance temporal robustness in training. Table~\ref{Tab:THUMOS14_Improve} depicts the performance gain of different TAD models trained using our method and tested with varying corruptions on the THUMOS14-C dataset. The robustness of different models has markedly improved,  even the model with the least improvement displays an absolute mAP rise of 2.36\% (averaged on 3 levels of 5 corruptions). 
When using the videoMAEv2 feature, the increase in robustness of ActionFormer is substantial. Specifically, both the mAP and relative robustness have an increase of 9.96\% and 13.22\% respectively. This indicates that our method can enhance the temporal anti-interference ability of the detection head, allowing it to better pair with VideoMAE and thereby achieve satisfactory results on clean data whilst maintaining commendable robustness. Table~\ref{Tab:Anet_Improve} displays the experimental results on the ActivityNet-v1.3-C dataset. It can be observed that, when trained with our method, the model is likewise capable of achieving consistent improvements in relative robustness across datasets with different levels of corruptions.

\begin{table}[t]
\tabcolsep 4pt
\centering
\caption{
Results of defending against temporal corruptions with the help of our proposed training strategy on ActivityNet-v1.3-C. Despite the challenges of this large-scale dataset, our method still yields consistent robustness improvements.}
\resizebox{\linewidth}{!}{
\begin{tabular}{l|l|l|l}
\hline
\multirow{2}{*}{Backbone(feature)} & \multirow{2}{*}{\shortstack{Clean \\ mAP}} & \multirow{2}{*}{\shortstack{Corrupted \\ mAP}}  & \multirow{2}{*}{\shortstack{Relative \\ Robutness}}\\ 
& & & \\
\hline
ActionFormer(TSP) & \textbf{36.50} & 27.79 & 76.12 \\
 + Ours & 36.01 (0.49 $\downarrow$) & \textbf{28.41 (0.62 $\uparrow$)} & \textbf{78.91 (2.79 $\uparrow$)} \\ \hline
ActionFormer(VideoMAEv2) & \textbf{38.47} & 33.93 & 88.19 \\
 + Ours & 38.44 (0.03 $\downarrow$) & \textbf{34.17 (0.24 $\uparrow$)} & \textbf{88.90 (0.71 $\uparrow$)} \\ \hline
TriDet(TSP) & \textbf{36.66} & 15.18 & 41.42 \\
 + Ours & 36.42 (0.24 $\downarrow$) & \textbf{17.28 (2.10 $\uparrow$)} & \textbf{47.45 (6.03 $\uparrow$)} \\ \hline
 AFSD(End-to-End) & 32.49 & 29.56 & 90.98 \\
 + Ours & \textbf{32.86 (0.37 $\uparrow$)} & \textbf{30.78 (1.22 $\uparrow$)} & \textbf{93.68 (2.70 $\uparrow$)} \\ \hline
\end{tabular}
}\label{Tab:Anet_Improve}
\end{table}

\begin{table}
\tabcolsep 3pt
    \centering
    \caption{Ablation study on our proposed training strategy, measured by the performance of TriDet on THUMOS14-C. Using FrameDrop and TRC loss simultaneously yields improvements in robustness and even mAP gain on clean videos.}
    \resizebox{\columnwidth}{!}{
    \begin{tabular}{cccccc}
    \hline
    \multirow{2}{*}{Feature} & \multirow{2}{*}{FrameDrop} & \multirow{2}{*}{TRC Loss} & \multicolumn{2}{c}{mAP (tIoU=0.5)} & \multirow{2}{*}{\begin{tabular}[c]{@{}c@{}}Relative \\ Robustness\end{tabular}} \\ \cline{4-5}
     &  &  & Clean & Corrupted &  \\ \hline
    \multirow{3}{*}{I3D} &  &  & 61.33 & 51.71 & 84.31 \\
     & \checkmark & \multicolumn{1}{l}{\textbf{}} & 62.37 & 52.56 & 84.27 \\
     & \checkmark & \checkmark & \textbf{62.63} & \textbf{54.07} & \textbf{86.33} \\ \hline
    \multirow{3}{*}{VideoMAEv2} &  &  & 75.16 & 61.10 & 81.29 \\
     & \checkmark & \multicolumn{1}{l}{\textbf{}} & 74.75 & 65.53 & 87.67 \\
     & \checkmark & \checkmark & \textbf{75.60} & \textbf{68.28} & \textbf{90.31} \\ \hline
    \end{tabular}
    }\label{Tab:ablation}
\end{table}


\subsection{Further Investigation of Our Training Strategy}\label{Sec:improvement}

\noindent \textbf{Effectiveness of our proposed training strategy}. 
Our proposed training strategy comprises two components: the FrameDrop strategy and the Temporal-Robust Consistency (TRC) loss. We conduct experiments by gradually adding them to the baseline. According to Table~\ref{Tab:ablation}, using only the FrameDrop strategy enhances the robustness of the TAD model across two types of features. However, when employing VideoMAEv2 features, the FrameDrop strategy alone does not improve the model's performance on clean videos. If both the FrameDrop strategy and TRC loss are utilized concurrently, not only is there a significant improvement in robustness, but the action detection performance on clean videos is also enhanced.

\noindent \textbf{Generality of our proposed training strategy}.
The length of the corruptions is fixed at one frame in our proposed training method. Our experiments show improvements on both datasets with the corruptions of varying lengths, suggesting that our method is applicable to corruptions of differing lengths. We also conduct experiments by training TAD models under one corruption while testing them on distinct corruptions. Table~\ref{Tab:generality} reveals that even if the corruptions are unseen during the training, our method still significantly enhances the model's robustness. This not only demonstrates the generality of our approach, unconstrained by any specific corruption but also indicates that our method does not simply train the model to memorize corruptions. Rather, it enhances the model's temporal robustness via our unique mechanism of disrupting temporal continuity and aligning localization distributions during the training. 

\begin{table}[t]
\tabcolsep 3pt
\centering
\caption{Performance comparison using different corruptions in the training, measured by the mAP of TriDet on THUMOS14-C. Our method is general and not limited to specific corruptions, and it consistently improves robustness on unseen corruptions.}
\resizebox{.75\columnwidth}{!}{
\begin{tabular}{l|c|c|c}
\hline
\diagbox[]{Test}{Train} & \multicolumn{1}{c|}{Clean} & \multicolumn{1}{c|}{Motion Blur} & Black Frame \\ \hline
Black Frame & \multicolumn{1}{c|}{43.47} & \multicolumn{1}{c|}{43.61 (0.14 $\uparrow$)} & 56.44 (12.97 $\uparrow$) \\ 
Packet Loss & \multicolumn{1}{c|}{63.54} & \multicolumn{1}{c|}{64.90 (1.36 $\uparrow$)} & 70.36 (6.82 $\uparrow$) \\
Overexposure & \multicolumn{1}{c|}{64.70} & \multicolumn{1}{c|}{64.77 (0.07 $\uparrow$)} & 70.01 (5.31 $\uparrow$) \\ 
Motion Blur & \multicolumn{1}{c|}{70.36} & \multicolumn{1}{c|}{73.62 (3.26 $\uparrow$)} & 74.84 (4.48 $\uparrow$) \\
Occlusion & \multicolumn{1}{c|}{63.44} & \multicolumn{1}{c|}{67.24 (3.80 $\uparrow$)} & 69.76 (6.32 $\uparrow$) \\ \hline
Average & \multicolumn{1}{c|}{61.10} & \multicolumn{1}{c|}{\textbf{62.83 (1.73 $\uparrow$)}} & {\textbf{68.28 (7.18 $\uparrow$)}} \\ \hline
\end{tabular} 
}
\vspace{-0.2cm}
\label{Tab:generality}
\end{table}

\section{Conclusion}
In this study, we have introduced a temporal robustness benchmark (THUMOS14-C and ActivityNet-v1.3-C), specifically designed for evaluating temporal action detection (TAD) methods. Unlike other video robustness benchmarks that apply corruptions to all frames of a video, we have designed a temporal corruption approach by corrupting a subset of frames within the video to disrupt its temporal continuity. We have conducted a robustness analysis on seven leading TAD methods, encompassing one-stage, two-stage, CNN-based and Transformer-based architectures. Our evaluation revealed that current TAD models are notably susceptible to temporal corruptions, with this vulnerability largely stemming from localization errors, rather than classification errors. We also observed that when corruption occurs in the middle of an action instance, TAD models tend to yield the largest performance drop.
These observations might suggest a promising direction for future research towards robust TAD. Furthermore, we have proposed a simple yet effective strategy for training temporally robust TAD models. This approach not only enhanced robustness but also improved performance on clean data. Given its universality and significance, robustness in TAD emerges as a new research dimension to be systematically explored in future studies.~\\

{
    \small
    \bibliographystyle{ieeenat_fullname}
    \bibliography{main,robust,acmart}
}

\clearpage
\setcounter{section}{0}
\renewcommand\thesection{\Alph{section}}
\setcounter{figure}{0}
\renewcommand\thefigure{\Alph{figure}}
\setcounter{table}{0}
\renewcommand\thetable{\Alph{table}}

\maketitlesupplementary


In the supplementary material, we provide more details and more experimental results of our work. We organize the supplementary into the following sections.
\begin{itemize}
    \item In Section~\ref{Sec:more_result}, we provide more ablated results of our Temporal-Robust Consistency (TRC) loss.
    \item In Section~\ref{Sec:more_DETAD}, we include more analysis using DETAD on the ActivityNet-v1.3-C and THUMOS14-C datasets. 
    \item In Section~\ref{Sec:examples}, we show more examples of videos corrupted by our proposed temporal corruptions. 
    \item In Section~\ref{Sec:specific}, we present detailed results of TAD models under the five types of corruptions on THUMOS14-C and ActivityNet-v1.3-C.
    \item In Section~\ref{Sec:action_background_training}, we investigated the difference between generating black frames in action-background pairs and solely within actions.
    \item In Section~\ref{Sec:multi-corruptions}, we present experiments involving the addition of various types of corruptions in action segments.
    \item In Section~\ref{Sec:more-type}, we demonstrate the experimental results with a wider range of corruptions types.
    \item In Section~\ref{Sec:more-dataset}, we provide the experimental results of adding temporal corruptions to the MultiThumos dataset.
\end{itemize}

\section{More Ablated Results of Our TRC Loss}\label{Sec:more_result}
\noindent \textbf{Action-cetric sampling strategy in TRC loss.}
As discussed in Section 5.2, considering the characteristics of TAD, we select predictions that are more temporally aligned with the action instance to compute the TRC loss. Here, we design two variants: \textbf{1) Full-Video}: using all predictions without sampling, and \textbf{2) Full-Action}: using predictions whose center falls within the action instance. As can be observed from Table \ref{Tab:sampling}, compared to the two variants, our proposed action-centric sampling method shows greater robustness on corrupted data and improvement on clean data while enjoying higher computational efficiency.
\begin{table}[!h]
\tabcolsep 3pt
\centering
\caption{Comparison of different sampling strategies in TRC loss, measured by the performance of TriDet on THUMOS14-C. Our action-centric sampling leads to the best results on both clean and corrupted data.}
\resizebox{0.7\columnwidth}{!}{
\begin{tabular}{c|c|c}
\hline
\multirow{2}{*}{Sampling Strategy} & \multirow{2}{*}{\shortstack{Clean \\ mAP}} & \multirow{2}{*}{\shortstack{Corrupted \\ mAP}} \\
 & & \\
\cline{1-3}
Without TRC & 75.16 &  61.10 \\
Full Video & 74.21 (0.95 $\downarrow$) & 67.21 (6.11 $\uparrow$) \\ 
Full Action & 75.04 (0.12 $\downarrow$) & 67.23 (6.13 $\uparrow$) \\
Action Center (Ours) & \textbf{75.60 (0.44 $\uparrow$)} &  \textbf{68.28 (7.18 $\uparrow$)} \\ 
\hline
\end{tabular}
}
\label{Tab:sampling}
\end{table}

\noindent \textbf{The Choice of Alignment Loss.}
Our approach to enhancing model robustness involves aligning predictions based on clean and corrupted videos. We compared Mean Square Error (MSE) and Kullback-Leibler (KL) divergence loss with our TRC loss. Models trained with different alignment losses are tested on clean and corrupted data. From Table \ref{Tab:loss}, all three types of losses can enhance the model's robustness on corrupted data, verifying the effectiveness of alignment. Notably, our proposed TRC loss not only enhances robustness but also improves the performance of clean data. Therefore, our method provides a new perspective that the performance and robustness of TAD methods can be simultaneously enhanced.

\begin{table}[!h]
\tabcolsep 3pt
\normalsize
\centering
\caption{Comparison of different alignment loss, measured by the performance of TriDet on THUMOS14-C. All losses improve robustness while only our TRC loss enhances the performance on clean data.}
\resizebox{0.7\columnwidth}{!}
{
    \begin{tabular}{l|c|c}
    \hline
    \multirow{2}{*}{Loss Function} & \multirow{2}{*}{\shortstack{Clean \\ mAP}} & \multirow{2}{*}{\shortstack{Corrupted \\ mAP}}  \\ 
     & & \\
    \cline{1-3}
    Without Alignment & 75.16 & 61.10 \\
    Mean Square Error & 74.59 (0.57 $\downarrow$) & {62.79 (1.69 $\uparrow$)} \\
    KL divergence & 73.32 (1.84 $\downarrow$) & {67.81 (6.71 $\uparrow$)} \\ 
    TRC (Ours) & \textbf{75.60 (0.44 $\uparrow$)} & {\textbf{68.28 (7.18 $\uparrow$)}} \\ 
    \hline
    \end{tabular}
}
\label{Tab:loss}
\end{table}

\begin{table*}[]
\tabcolsep 3pt
\centering
\begin{minipage}{\textwidth}
    \caption{The performance of TAD models concerning corruption robustness on THUMOS14-C, considering five distinct types of corruptions and three different levels, measured by mAP when the tIoU is set to 0.5.}
    \resizebox{\textwidth}{!}{%
    \begin{tabular}{cc|c|ccc|ccc|ccc|ccc|ccc}
        \hline
         &  &  & \multicolumn{15}{c}{Corruption Type}\\
        \cline{4-18}
        Model& Feature & Clean Frame& \multicolumn{3}{c|}{Black Frame} & \multicolumn{3}{c|}{Packet Loss} & \multicolumn{3}{c|}{Overexposure} & \multicolumn{3}{c|}{Motion Blur} & \multicolumn{3}{c}{Occlusion} \\
        \cline{4-18}
        & & & Level 1 & Level 2 & Level 3 &  Level 1& Level 2 & Level 3 &  Level 1& Level 2 & Level 3 &  Level 1& Level 2 & Level 3 &  Level 1& Level 2 & Level 3 \\
        \hline
         BasicTAD &  SlowOnly & 59.17 & 46.82& 26.07 & 16.23 & 56.17 & 45.64 & 41.53 & 54.78 & 33.63 & 21.95 & 53.85 & 36.98 & 28.51 & 48.66 & 32.41 & 22.54\\
        
         E2E-TAD & SlowFast & 56.41 & 29.23 & 15.54 & 12.22 & 40.32 & 32.07 & 30.23 & 50.03 & 24.33 & 14.13 & 45.33 & 41.51 & 39.30 & 38.86 & 25.73 & 19.46\\
        
        TemporalMaxer  & I3D & 60.72 & 52.83 & 40.12 & 23.53 & 58.71 & 56.51 & 52.62 & 51.71 & 44.83 & 36.28 & 57.79 & 52.40 & 42.97 & 56.01 & 49.17 & 41.88\\
        
        ActionFormer & I3D & 61.53 & 54.89 & 44.53 & 29.43 & 58.96 & 57.39 & 55.13 & 54.14 & 47.84 & 41.01 & 58.89 & 54.09 & 46.36 & 57.99 & 52.07 & 46.43\\
        
        ActionFormer &  VideoMAEv2 & 73.84 & 62.83 & 41.19 & 22.34 & 68.01 & 61.91 & 56.43 & 67.90 & 62.09 & 58.46 & 71.13 & 68.88 & 66.30 & 64.57 & 54.56 & 48.30\\
        
        AFSD &  I3D & 46.05 & 38.69 & 28.70 & 22.12 & 43.28 & 39.37 & 36.79 & 39.02 & 29.57 & 23.40 & 41.65 & 34.15 & 27.45 & 42.37 & 37.49 & 33.00\\
        
        TriDet &  I3D & 61.33 & 55.61 & 46.74 & 33.08 & 59.94 & 57.95 & 55.90 & 54.63 & 48.94 & 43.35 & 59.30 & 54.20 & 47.65 & 58.56 & 52.90 & 46.88\\
        
        TriDet &  VideoMAEv2 & 75.16 & 64.39 &42.42 & 23.60 & 68.99 & 64.26 & 57.39 & 69.83 & 64.04 & 60.24 & 72.53 & 70.72 & 67.81 & 69.32 & 62.92 & 58.07\\
        \hline
    \end{tabular}}
    \label{Tab:THUMOS14-C}

    \vspace{20pt}

    \caption{The performance of TAD models concerning corruption robustness on ActivityNet-v1.3-C, considering five distinct types of corruptions and three different levels, measured by the average mAP of the tIoU thresholds between 0.5 and 0.95 with the step of 0.05.}
    \resizebox{\textwidth}{!}{%
        \begin{tabular}{cc|c|ccc|ccc|ccc|ccc|ccc}
        \hline
         &  & & \multicolumn{15}{c}{Corruption Type}\\
        \cline{4-18}
        Model& Feature& Clean Frame & \multicolumn{3}{c|}{Black Frame} & \multicolumn{3}{c|}{Packet Loss} & \multicolumn{3}{c|}{Overexposure} & \multicolumn{3}{c|}{Motion Blur} & \multicolumn{3}{c}{Occlusion} \\
        \cline{4-18}
        &  & & Level 1 & Level 2 & Level 3 &  Level 1& Level 2 & Level 3 &  Level 1& Level 2 & Level 3 &  Level 1& Level 2 & Level 3 &  Level 1& Level 2 & Level 3 \\
        \hline
         VSGN & I3D & 31.85 & 29.72 & 28.04 & 24.44 & 31.73 & 31.08 & 30.69 & 31.60 & 30.13 & 28.35 & 31.72 & 30.87 & 29.82 & 31.72 & 31.04 & 30.19\\
    
         TriDet & TSP & 36.66 & 22.23 & 10.69 & 6.66 & 20.58 & 13.91 & 12.82 & 20.70 & 13.21 & 11.79 & 20.67 & 13.94 & 12.77 & 20.59 & 14.16 & 13.04\\
        
        ActionFormer  & TSP & 36.50 & 30.59 & 20.34 & 8.44 & 30.17 & 29.92 & 29.84 & 29.72 & 29.12 & 28.57 & 30.19 & 29.83 & 29.64 & 30.23 & 30.12 & 30.06\\
        
        ActionFormer & VideoMAEv2 & 38.47 & 37.01 & 14.19 & 7.37 & 38.46 & 38.26 & 38.08 & 38.28 & 37.37 & 36.41 & 38.36 & 37.44 & 36.32 & 38.13 & 37.04 & 36.16\\
        AFSD &  I3D & 32.49 &  30.19 & 24.73 & 19.36 & 32.13 & 31.22 & 30.43 & 31.28 & 29.07 & 27.69 &32.08 & 30.74 & 29.50 & 32.21 & 31.52 & 31.20\\
        \hline
        \end{tabular}
    }
    \label{Tab:Anet-C}
\end{minipage}
\end{table*}

\begin{figure*}[!t]
    \centering
    \begin{minipage}{.47\textwidth}
        \includegraphics[width=\columnwidth]{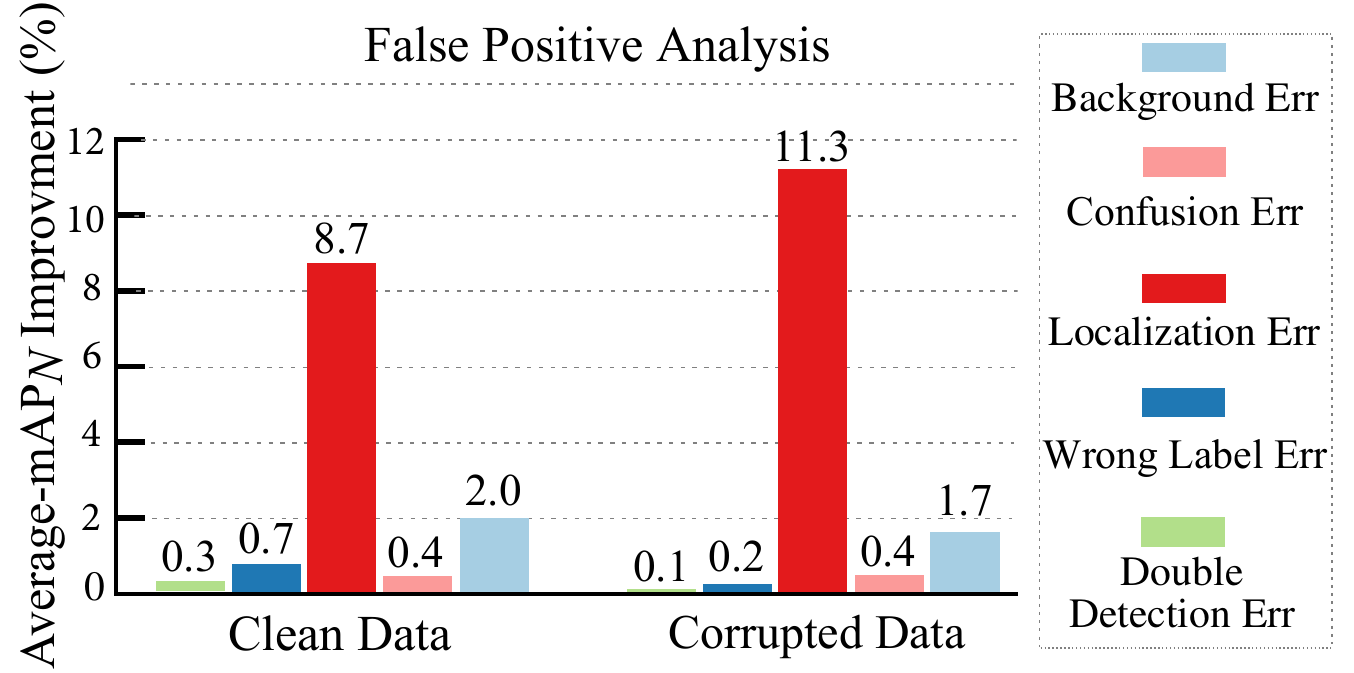}
        \caption{False positive profiling of the TriDet's predictions on ActivityNet-v1.3-C.  The Wrong Label (classification) Error remains relatively consistent, whereas the Localization Error increases significantly on corrupted data, revealing that vulnerability mainly comes from localization error rather than classification error.}
        \label{Fig.detad_tridet_anet}
    \end{minipage}
    \hspace{0.52cm}
    \begin{minipage}{.47\textwidth}
        \includegraphics[width=\columnwidth]{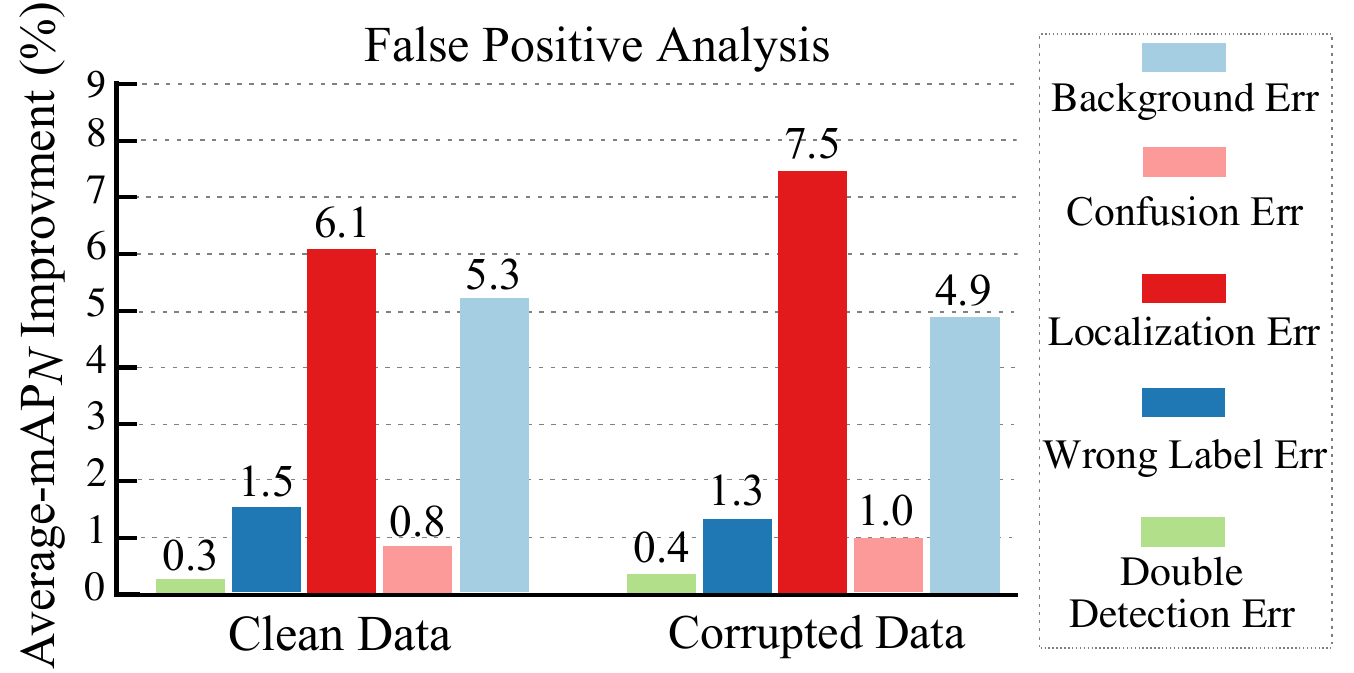}
        \caption{False positive profiling of the ActionFormer's predictions on THUMOS14-C. The Localization Error of corrupted data is notably higher than that with clean data.}
        \label{Fig.detad_af_thumos} 
    \end{minipage}
\end{figure*}

\section{More Analysis Using DETAD~\cite{alwassel2018diagnosing}}\label{Sec:more_DETAD}
\noindent We present further analyses for more TAD models and more datasets using DETAD, a tool for diagnosing TAD models. Figure \ref{Fig.detad_tridet_anet} depicts the results of analysis using DETAD on the predictions of TriDet on ActivityNet-v1.3-C, while Figure \ref{Fig.detad_af_thumos} illustrates the analysis of ActionFormer on THUMOS14-C. Evidently, the conclusions align consistently with those presented in the main paper across different models and datasets. Specifically, on our corrupted datasets, the results indicate a significant increase in localization errors, with minimal change in classification errors. This observation underscores the critical corruption introduced by our dataset, emphasizing its impact on temporal continuity (\ie, localization) rather than compromising action recognition (\ie, classification).


\section{More Examples of Corrupted Videos}\label{Sec:examples}
\begin{figure*}[!t]
    \centering
    \includegraphics[width=0.78\textwidth]{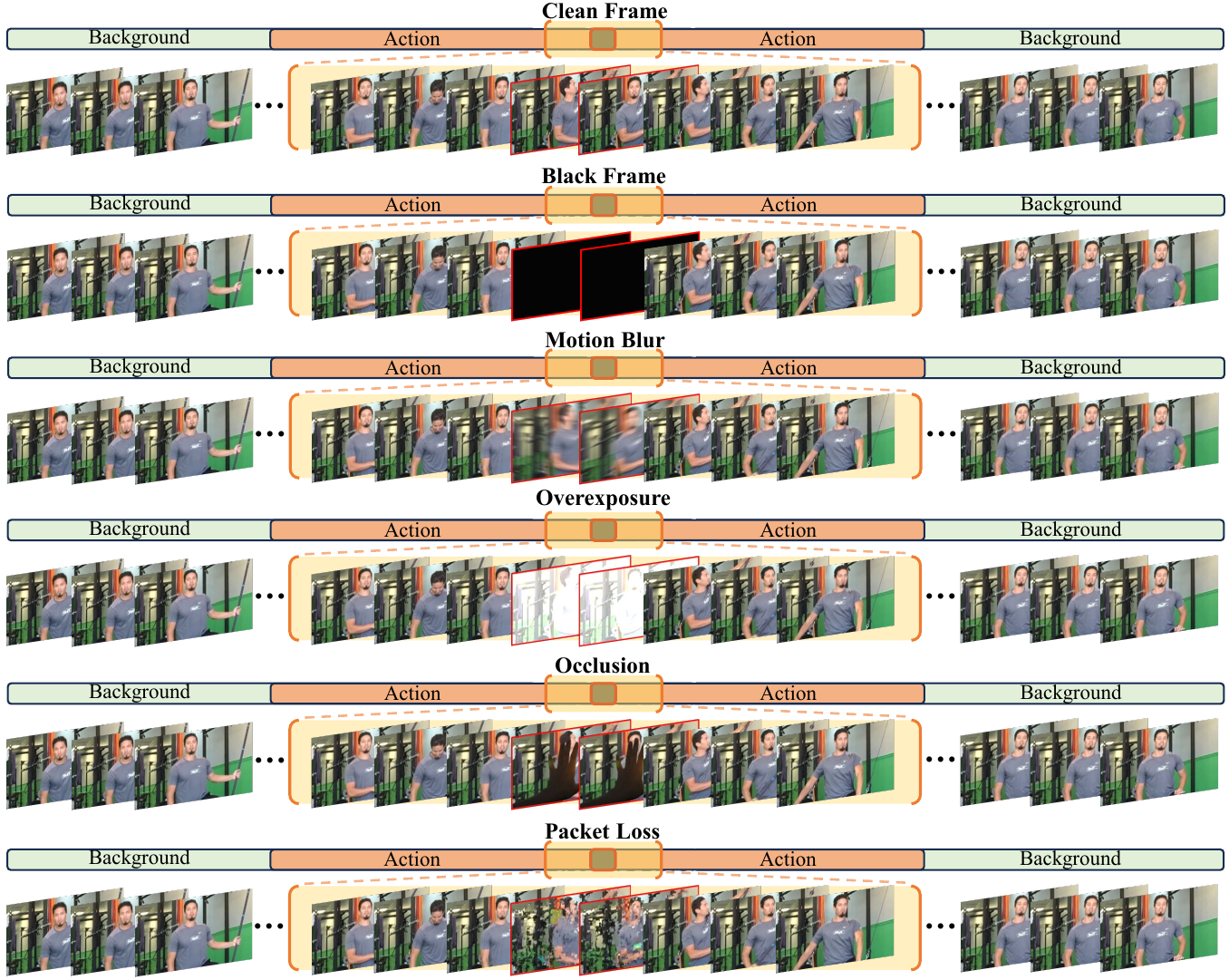}
    \centering
    \includegraphics[width=0.78\textwidth]{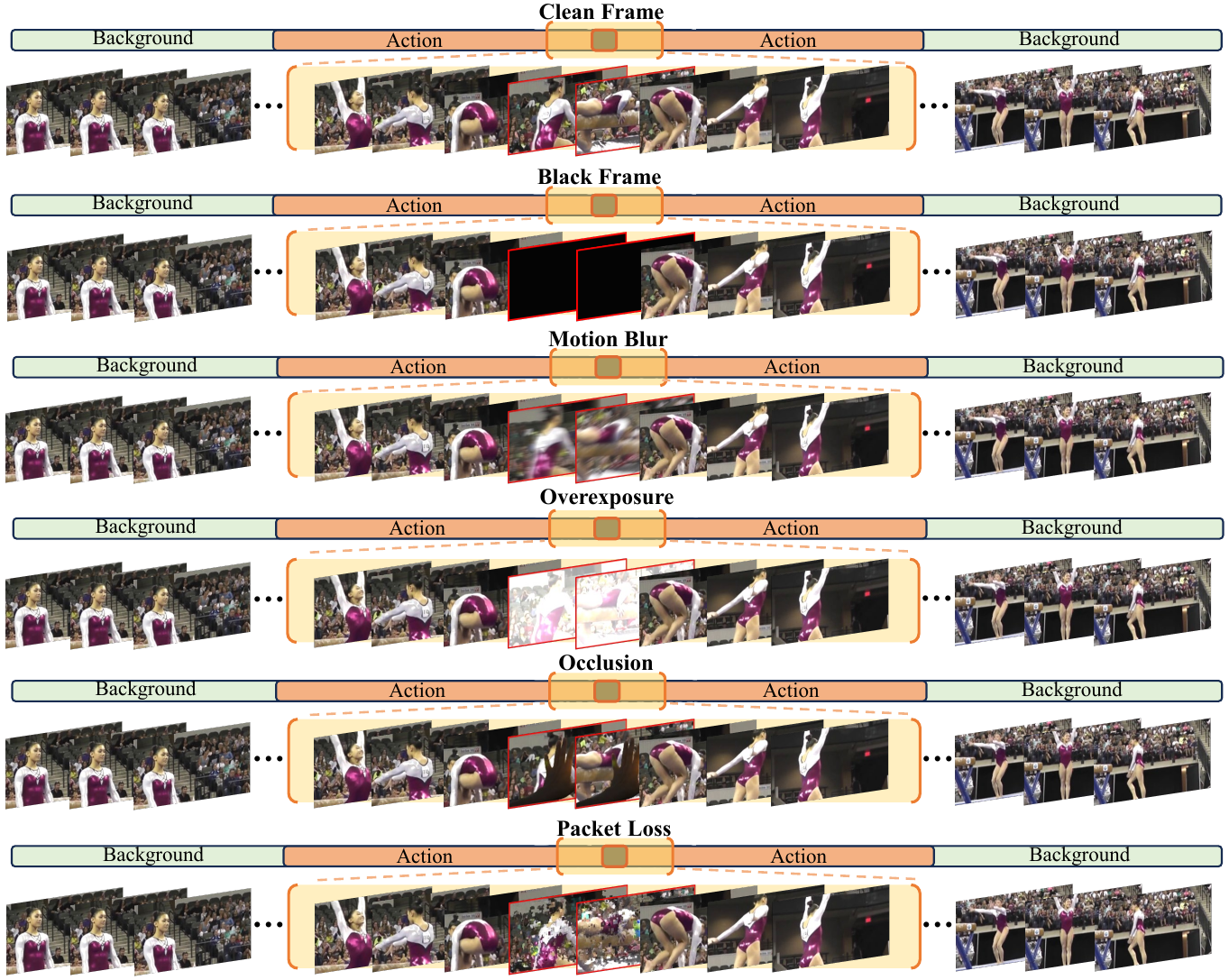}
    \caption{More examples of our temporal corruptions dataset.}
    \label{Fig.example} 
\end{figure*}

\noindent Figures~\ref{Fig.example} present more illustrative examples demonstrating the incorporation of five distinct types of corruptions into the actions depicted in the video. Although we only corrupt a small portion of frames within the action, and some of these corruptions may not appear significantly different from the surrounding clean frames—at least not to the extent of notably hindering human ability to locate actions—we discover through our experiments that even such subtle corruptions can substantially impair the localization capability of Temporal Action Detection (TAD) models. This finding indicates that the type of corruption we have introduced serves as an effective means to evaluate the temporal robustness of TAD models.

\section{Performance of TAD models udner Each Type/Level of Temporal  Corruptions}\label{Sec:specific}
\noindent We present the detailed performance of different TAD models under the temporal corruptions of different types and levels in Tables~\ref{Tab:THUMOS14-C} and~\ref{Tab:Anet-C}. It reads that existing TAD models are particularly vulnerable to temporal corruptions.

\section{Comparisons on different black-frame locations for training.}\label{Sec:action_background_training}
\noindent As discussed in Section 5.1, our proposed FrameDrop Strategy generates black frames in action-background pairs. We also attempted to generate black frames solely in actions for model training. Employing these two methods on the THUMOS14-C dataset to train TemporalMaxer model, we obtained the results shown in Table~\ref{tab:corrupt_in_action}. It is evident that while the method of generating black frames solely in actions enhances the model's robustness, its performance on clean data diminishes. The reason may be that the model learns a bias—memorizing that corruptions is expected to occur within the action. Thus, we opt for corrupting the action-background pair in the FrameDrop Strategy.

\begin{table}[h  ]
\tabcolsep 2.5pt
  \centering
  \caption{Comparisons on different black-frame locations. It is clear that while the approach of exclusively generating black frames within actions improves the model's robustness, it leads to a decrease in performance on clean data.}
  \resizebox{\columnwidth}{!}
 {
    \begin{tabular}{l|l|cccc}
        \hline
         THUMOS14-C & {\diagbox[height=2em, width=5em]{Test}{Train}} & Clean & Action & Action-background   \\
        \hline
         
    \multirow{2}{*}{TemporalMaxer~[68]} & Clean & 60.72 & \textcolor{red}{59.28(\textbf{1.44 $\downarrow$})}  & \textbf{61.04 (0.32 $\uparrow$)}      \\
     & Corrupted &  47.82       &    53.64(5.82 $\uparrow$)    &   51.95(4.13 $\uparrow$)     \\
    \hline
    \end{tabular}%
    }
    \label{tab:corrupt_in_action}
\end{table}%

\section{Multiple types of corruptions.}\label{Sec:multi-corruptions}
\noindent Building upon the addition of the five types of corruption mentioned in the main text, we also conducted two experiments. In these experiments, we randomly selected two corruption types and then: 1) applied both types to all middle frames (\textbf{spatial}) or 2) divided the middle frames temporally and applied one type to each half (\textbf{temporal}). We compared the results of TriDet and TemporalMaxer models on the THUMOS14-C dataset without adding corruptions (\textbf{clean}), adding only one type of corruptions (\textbf{ours}), and adding multiple types of corruptions simultaneously, as shown in Table~\ref{tab:multi_corrupt}. The experimental results demonstrate that the simultaneous presence of multiple corruptions often degrades model robustness more significantly than the presence of only one type of corruptions.

\begin{table}[h]
\tabcolsep 2pt
  \centering
  \caption{Compared the performance of the TriDet and TemporalMaxer models on the THUMOS14-C dataset.}
 \resizebox{\columnwidth}{!}
 {
    \begin{tabular}{lcccc}
        \hline
          Corruption (\# types) & Clean (0) & Ours (1) & Spatial (2) & Temporal (2) \\
        \hline
    TriDet [60] & 61.33 & 51.71 & 43.31 & 50.85 \\
    TemporalMaxer [68] & 60.72 & 47.82 & 40.84 & 49.43 \\
    \hline
    \end{tabular}%
    }
    \label{tab:multi_corrupt}
\end{table}%

\section{More types of corruption.}\label{Sec:more-type}
\noindent In addition to the five types of corruptions mentioned in the main text, we also experimented with the effects of four additional types of corruptions, including: 
\begin{itemize}
    \item \textbf{Jittering}: caused by camera shake during filming
    \item \textbf{Different frame rate}: resulting from bandwidth limitations during network transmission
    \item \textbf{Slow-motion}: common shot types in videos
    \item \textbf{Time-lapse}: arising from video buffering issues and limitations in the processing power of playback devices
\end{itemize}

We tested TemporalMaxer model on the THUMOS14-C dataset, and the experimental results are shown in Table~\ref{tab:4corruption}. It can be seen that these four additional types of corruptions also significantly degrade the model's performance. This indicates that the degradation of model performance due to corruptions is independent of the corruptions type, suggesting that any corruptions in real-world scenarios could potentially lead to a decrease in model performance.

\begin{table}[h]
\tabcolsep 1.5pt
  \centering
  \caption{The average mAP of TemporalMaxer model with the addition of four types of corruptions at each corruptions level on the THUMOS14-C dataset. The experimental results indicate that these four types of corruptions also lead to a decrease in model performance.}
  \resizebox{\columnwidth}{!}
 {
    \begin{tabular}{lccccc}
        \hline
          Corruption  & Clean & Jittering & Frame rate & Slow-motion & Time-lapse \\
    \hline
    TemporalMaxer~[68] &    60.72     &   50.16    &   40.05    &   56.03    & 45.25  \\
    \hline
    \end{tabular}%
    }
    \label{tab:4corruption}
\end{table}%

\section{More datasets for general evaluation.}\label{Sec:more-dataset}
\noindent In addition to conducting experiments on the commonly used THUMOS14 and ActivityNet-v1.3 datasets, we also attempted to construct MultiThumos-C using the same method as a benchmark for testing model robustness on the MultiThumos dataset. We tested this benchmark using the TemporalMaxer model and obtained the average mAP under five types of corruptions, as shown in Table~\ref{tab:multithumos}. The results indicate that adding corruptions to this dataset also significantly degrades model performance.
\begin{table}[h]
\tabcolsep 2pt
  \centering
  \caption{The average mAP of TemporalMaxer on the MultiThumos-C dataset for each type of corruptions. The results indicate that adding corruptions to the MultiThumos dataset also significantly degrades model performance.}
  \resizebox{\columnwidth}{!}
 {
    \begin{tabular}{l|ccc|c}
        \hline
          tIoU & 0.2 & 0.5 & 0.7 & Average  \\
    \hline
    Clean &     44.30   &   30.54    &   15.72    &   27.55   \\
    Corrupted  &     41.24  (\textbf{3.06 $\downarrow$})   &   26.67 (\textbf{3.87 $\downarrow$})   &   12.65  (\textbf{3.07 $\downarrow$})  &   24.70 (\textbf{2.85 $\downarrow$})  \\
    \hline
    \end{tabular}%
    }
    \label{tab:multithumos}
\end{table}%

\end{document}